\documentclass[conference]{IEEEtran}
\IEEEoverridecommandlockouts
\usepackage{cite}
\usepackage{amsmath,amssymb,amsfonts}
\usepackage[ruled,vlined]{algorithm2e}

\usepackage{url}
\usepackage{array}
\usepackage{tabularx}
\usepackage{algpseudocode}
\usepackage{booktabs}
\usepackage{subcaption}
\usepackage{multirow}
\usepackage{graphicx}
\usepackage{textcomp}
\usepackage{xcolor}
\usepackage{xspace}
\usepackage[numbers,sort&compress]{natbib}
\def\BibTeX{{\rm B\kern-.05em{\sc i\kern-.025em b}\kern-.08em
    T\kern-.1667em\lower.7ex\hbox{E}\kern-.125emX}}

\newcommand{\model}{RAFG\xspace}
\newcommand{\modelbf}{\textbf{RAFG}\xspace}

\definecolor{bgcolor1}{HTML}{DAE8FC}
\definecolor{bgcolor2}{HTML}{D5E8D4}
\definecolor{bgcolor3}{HTML}{E1D5E7}

\begin{document}

\title{Retrieval-Augmented Feature Generation for Domain-Specific Classification
 }


\author{
    \IEEEauthorblockN{
        Xinhao Zhang\textsuperscript{1*}\thanks{* These authors contributed equally to this work.},
        Jinghan Zhang\textsuperscript{2*},
        Fengran Mo\textsuperscript{3},
        Dakshak Keerthi Chandra\textsuperscript{4},\\
        Yu-Zhong Chen\textsuperscript{$\lozenge$}\thanks{$\lozenge$ Independent researcher.},
        Fei Xie\textsuperscript{1},
        Kunpeng Liu\textsuperscript{2†}\thanks{† Corresponding author.}
    }
    \IEEEauthorblockA{
        \textit{\textsuperscript{1}Portland State University, USA}, 
        \textit{\textsuperscript{2}Clemson University, USA}, 
        \textit{\textsuperscript{3}University of Montreal, Canada}, 
        \textit{\textsuperscript{4}Analog Devices, USA} \\
        \{xinhaoz, xie\}@pdx.edu,
        \{jinghaz, kunpenl\}@clemson.edu,
        fengran.mo@umontreal.ca,\\
        dakshak.keerthichandra@analog.com,
        yuzhong.chen@asu.edu
    }
}

\maketitle

\begin{abstract}
Feature generation can significantly enhance learning outcomes, particularly for tasks with limited data. An effective way to improve feature generation is to expand the current feature space using existing features and enriching the informational content. However, generating new, interpretable features usually requires domain-specific knowledge on top of the existing features. 
In this paper, we introduce a \underline{R}etrieval-\underline{A}ugmented \underline{F}eature \underline{G}eneration method, \model, to generate useful and explainable features specific to domain classification tasks. 
To increase the interpretability of the generated features, we conduct knowledge retrieval among the existing features in the domain to identify potential feature associations. These associations are expected to help generate useful features.
Moreover, we develop a framework based on large language models (LLMs) for feature generation with reasoning to verify the quality of the features during their generation process. Experiments across several datasets in medical, economic, and geographic domains show that our \model method can produce high-quality, meaningful features and significantly improve classification performance compared with baseline methods.
\end{abstract}


\section{Introduction}

In domain-specific applications, e.g., disease classification and insurance claim prediction, the data are usually scarce and result in limited features~\cite{kupinski1999feature,jain2018feature,taha2022using}. In these cases, it is infeasible to train a machine learning model with satisfactory performance.
A common practice is to generate new features to support the model training. However, manual feature generation relies heavily on domain expertise, while automated methods might produce features that lack interpretability~\cite{zhang2023interpretable,kuznetsov2021interpretable}.
Existing studies typically focus solely on optimizing data structures~\cite{guo2017deepfm}. Although efficient, these approaches lack transparency and interpretability.
In addition, obtaining the resources of domain-specific knowledge is expensive~\cite{tricot2014domain,wang2023human}.

\begin{figure*}[ht]
    \centering
    \includegraphics[width=0.98\textwidth]{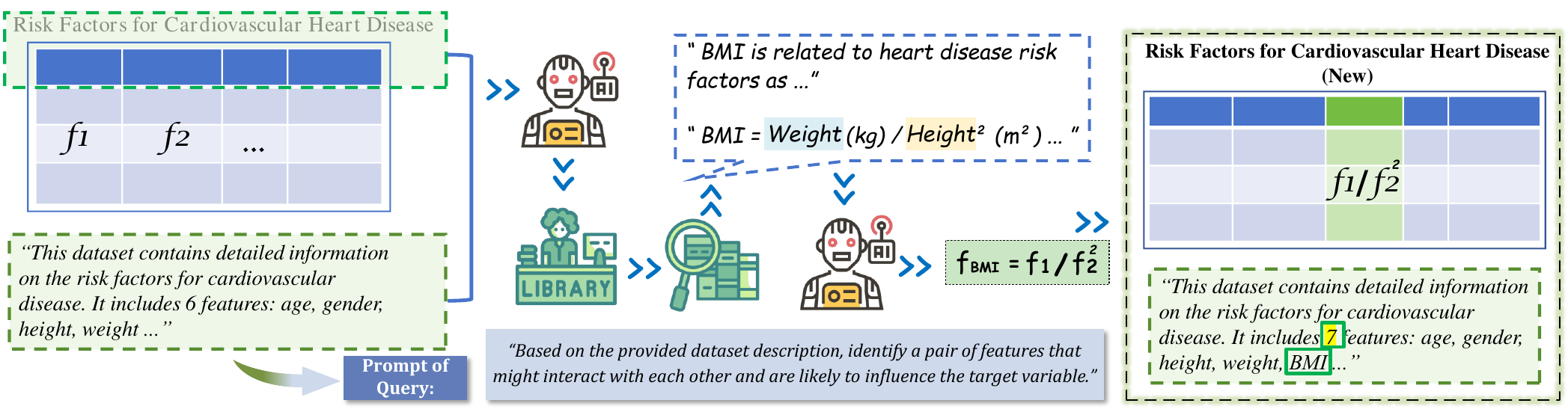}
    \caption{Framework of \model. We adopt an LLM to generate new features according to the retrieved textual information containing expertise knowledge (e.g., the BMI as shown case).}
    \label{fig:intro}
\end{figure*}

An intuitive solution is to utilize the knowledge among existing features and their associations by extracting and understanding their underlying relationships. This process is non-trivial. It requires referencing domain knowledge to generate new features, a challenging task for the models without specific fine-tuning. 
Although fine-tuning a language model with domain-specific data can inject domain knowledge, it is costly and might limit a model's generalizability across different domains~\cite{zhang2021domain,zhang2022learning}.
Thus, an alternative is to retrieve useful information from an external knowledge base as supporting references. Then,  a generator model with a reasoning mechanism, e.g., a Large Language Model (LLM), can be used to integrate the original feature with the corresponding support knowledge to generate the new feature. 
This is similar to the retrieval-augmented generation (RAG)~\cite{lewis2020retrieval,fan2024survey,yu2024auto} technique, where we inherit a similar principle for domain-specific feature generation. 
On the one hand, the retrieved knowledge is expected to serve as explicit references for interpretability~\cite{jeong2024llm,han2024large}. On the other hand, the new features are generated based on the existing features, which can reduce the representation inconsistent risk and leverage limited data efficiently~\cite{wang2024controllable,islam2022comprehensive}. The crucial question is \textit{how to incorporate the external retrieved information with the existing feature to support useful feature generation?}

We depict a comprehensive example in Fig.~\ref{fig:intro} for the feature generation task. The goal is to detect heart disease based on the generated features. 
To solve the task, a more concrete indicator is the \textit{BMI} feature, which could be generated by the original features \textit{Height} and \textit{Weight}. 
This desirable information can be generated with the support of external knowledge of \textit{BMI} via retrieval. 
Such insights serve as our motivation for implementing retrieval-augmented feature generation. 

\noindent \textbf{Our Targets.} We aim to address three main challenges in generating features for domain-specific scenarios: 1) how to extract and utilize useful information by understanding their relations, 2) how to apply the retrieval-augmented generation paradigm to retrieve useful knowledge with LLMs' reasoning capability for generating informative and explainable domain features, and 3) how to form a high quality feature space on top of the original and generated features.

\noindent \textbf{Our Approach.} To address these challenges, we propose an LLM-based \textbf{R}etrieval-\textbf{A}ugmented \textbf{F}eature \textbf{G}eneration approach (\model) to generate features with retrieved knowledge in terms of the existing features. Specifically, we first instruct an LLM to analyze and capture the implicit pattern within the structured data and the textual information, e.g., dataset descriptions and feature labels. 
With advanced reasoning capabilities, the LLM is expected to provide a logical process to derive the relation among the features and generate a query for later domain-specific retrieval. Then, we deploy the RAG technique to identify the useful references from a reliable knowledge base according to the previously generated query.
The used original features, together with the retrieved knowledge, are fed to the LLM to generate new features according to the reasoning capacity of LLM.
Eventually, an automated feature generation mechanism that consistently refines the generated feature and is integrated into the final data attribution. This is supported by an iterative feature validation mechanism to update the feature space with validated ones.
The iteratively validated features via generation are expected to enhance the domain-specific interpretability by aligning with the needs of the fields and supply additional crucial indicators. 

Our contributions are threefold:
\begin{enumerate}
    \item We introduce a novel LLM-based retrieval-augmented feature generation method for domain-specific scenarios, which extracts and utilizes the potential relations among the existing features for new feature generation. 
    \item The \model framework is training-free and can be adapted for any specific domains with an automatic validation mechanism. 
    The generated feature is more explainable and enhances the domain-specific classification performance with the retrieved reference.    
    \item Experimental results on domain-specific downstream tasks demonstrate the effectiveness of our \model by outperforming other feature extraction approaches.

\end{enumerate}
\section{Related Work}

\subsection{Feature Generation}
Feature engineering is an essential part of machine learning, including feature selection, feature editing, or generating new features from raw data to improve the model performance~\cite{nargesian2017learning,dong2018feature,salau2019feature}. The target of feature engineering is to optimize the feature representation space to benefit the downstream tasks~\cite{chandrashekar2014survey,zozoulenko2025random,zhang2024dynamic}. 
Feature generation is an effective approach in the context of feature engineering, which aims to create complex latent feature spaces~\cite{yang2024graphusion} through various mathematical or logical transformation operations~\cite{guo2017deepfm,xiao2024traceable,zhang2019job2vec}. 
The advancement of deep learning techniques and LLMs provides new opportunities to enhance the transparency and alleviate the manual operations~\cite{pan2020deep,bjorneld2024real} for feature generation.
Existing studies have proposed various methods, ranging from traditional techniques~\cite{hassan2023comparative} to more sophisticated ones. For instance, Cognito~\cite{khurana2016cognito} is a reinforcement learning approach for intelligent feature space exploration~\cite{khurana2018feature}. Besides, graph-based feature construction~\cite{zhang2023openfe}, policy networks for pre-training feature engineering~\cite{li2023learning}, and multi-stage approaches combining statistical selection with neural generation~\cite{horn2019autofeat,dhal2022comprehensive} are developed. 
Though showing efficiency, these methods 
still focus on structural and numerical data, which might overlook semantic information and lack interpretability. 
More recently, CAAFE~\cite{hollmann2024large} leverages LLMs to generate context-aware features with explanations~\cite{gong2025evolutionary,wang2025diversity,zhang2024prototypical,galindo2023large,yu2023large} to enhance the interpretability. 
Different from them, we propose a retrieval-augmented feature generation framework to facilitate the model's prediction in the downstream tasks with better performance and interpretability.

\subsection{Retrieval-Augmented Machine Learning}
Retrieval-Augmented Generation (RAG)~\cite{lewis2020retrieval} aims to retrieve external knowledge to enhance the capacity of language models.
The retrieved knowledge is used according to the requirements of downstream tasks~\cite{hu2024rag}, such as reducing hallucination~\cite{shuster2021retrieval,yang2024leandojo}, supplying up-to-date knowledge for accurate answers~\cite{sanmartin2024kg,zhang2024raft,mo2025uniconv}, and assisting in developing better machine learning models via generating useful features~\cite{zamani2022retrieval,shi2025flexolmo}.
However, the retrieved knowledge could not be directly used for feature generation without further understanding and reasoning by LLMs~\cite{kaul2017autolearn,zhang2023openfe,wang2025diversity}.
To generate interpretable features, the LLM needs to analyze the relevance and rationality of the retrieved information and the generated features~\cite{han2023comprehensive,zhang2025leka,mo2025convmix}. 
Then, the reasoning process is expected to enhance the new features by validating their effectiveness and explainability in the domain knowledge, which is the focus of our studies.

\section{Problem Statement}
 
We formulate the task as generating new optimal and explainable feature representations via searching external knowledge on top of the existing feature.
Concretely, we denote the original tabular dataset as $\mathcal{D}_0=\{ \mathcal{F}_0;\mathit{y}\}$ that includes an original feature set $\mathcal{F}=\{f_i\}_{i=1}^{I_0}$, each features $\{f_i\}$ and its target label $\mathit{y}$ as well as textual information $\mathcal{C}_0$ including the labels and the data description. Our optimization objective is to automatically generate a set of new features $\{ g_{t}\}$ based on the retrieved knowledge with the reasoning procedure supported by an LLM that can reconstruct an optimal feature set $\mathcal{F}^*$ as
\begin{equation}
    \mathcal{F}^* = \underset{\hat{\mathcal{F}}}{\text{argmax}} \; \mathbf{\mathcal{P}}_{\mathcal{A}}(\hat{\mathcal{F}}, \mathit{y}),
\end{equation}

\noindent where $\mathcal{A}$ is a domain-specific downstream task (e.g., predicting Parkinson’s disease), $\mathcal{P}$ is the performance indicator of $\mathcal{A}$ and $\hat{\mathcal{F}}=\{\mathcal{F}_0, \{g_{t}\} \}$ is an optimized feature set. 

\section{Methodology}

\begin{figure*}[ht]
    \centering
    \includegraphics[width=\textwidth]{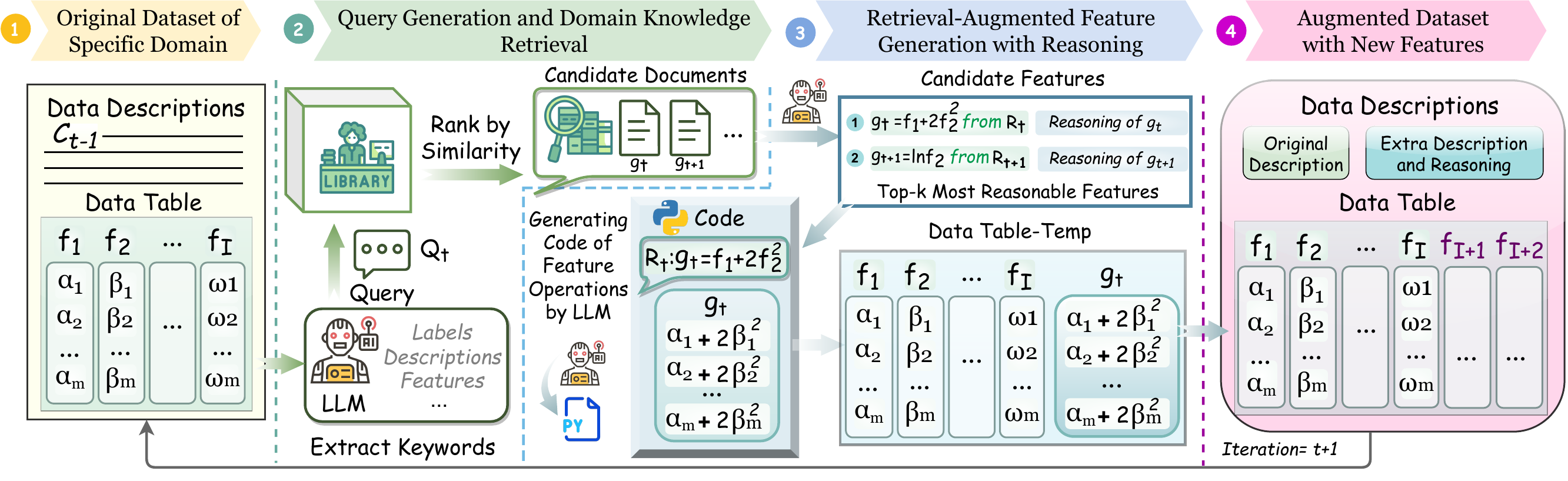}
    \caption{Framework of \model. Given an input data table including a description, feature vectors, and a target label vector, the LLM first integrates the text information of description, label information, and data types to embed and form a query. Then, with this query we adopt RAG technology to search through an external library for one of several relevant documents which can guide the LLM in creating a new feature with most potential. After that, we test the template data table with the new feature for metrics improvement, and the LLM decides whether to reserve this new feature. This searching and generation process iterates until reaching the maximum rounds of iteration, or the best feature space is found.}
    \label{fig:method}
    \vspace{-3mm}
\end{figure*}
In this section, we introduce our novel feature generation method \underline{\textbf{R}}etrieval-\underline{\textbf{A}}ugmented \underline{\textbf{F}}eature \underline{\textbf{G}}eneration (\modelbf), which generates features automatically by the retrieved external knowledge and the original text information. 
We utilize an LLM as a text-miner and a feature generator, which generates the query for retrieval and integrates the search results with its parametric for feature generation.
An automatic mechanism is applied to validate the generated features and the original ones in the feature set, enabling continuous optimization and updating of the feature content.
An overview of the \modelbf framework is shown in Fig.~\ref{fig:method}, including three stages: \textit{A. Query Generation and Domain Knowledge Retrieval; B. Retrieval-Augmented Feature Generation with Reasoning}; and \textit{C. Feature Content Validation by LLM.}

\subsection{Query Generation and Domain Knowledge Retrieval}
We aim to retrieve the external knowledge on top of the original features to augment the information for new feature generation interactively.
To this end, given the $t$-th iteration, we first collect the textual information of the current feature set $\mathcal{F}=\mathcal{F}_{t-1}=\{f_i\}_{i=1}^{I}$, including the goals of the downstream tasks $\mathcal{A^{\prime}}$, and the textual information $\mathcal{C}=\mathcal{C}_{t-1}$ to construct a query $\mathcal{Q}_t$ via an LLM $p_\phi$ as
\begin{equation}
    \mathcal{Q}_t = p_\phi(\mathcal{A}^{\prime}, \mathcal{C}_{t-1}).
\end{equation}
We then retrieve the top-$k$ relevant documents from the whole external knowledge base in terms of the query $\mathcal{Q}_t$. 
The relevance between each candidate document and the query is determined by a computation of cosine similarity as 
\begin{equation}
    \text{sim}(\mathcal{Q}_t, r_j) = \frac{\mathcal{Q}_t \cdot r_j}{\|\mathcal{Q}_t\| \|r_j\|} ,
\end{equation}
where $r_j$ is the candidate document and $r_j \in \mathcal{R}_k$.
Then, the set of retrieved documents $\{\mathcal{R}_k\}$ are considered to consist of a set of potential features $\{g_{k}\}$ further determined by LLM to update the previous feature set $\mathcal{F}_{t-1}$ as
\begin{equation}
    g_k = \{p_\phi (\mathit{o}(\mathcal{F}_{t-1}) \mid \mathcal{R}_k )\}, 
\end{equation}
\noindent where $\mathit{o}$ denotes the various operations based on LLM, which are denoted as follows:
\begin{itemize}
    \item \textit{Scaling}: This operation involves arithmetic transformations, statistical measures, or algorithmic functions applied to derive $g_t$, i.e., a mapping function $\mathit{o}_s: \mathbb{R}^n \rightarrow \mathbb{R}^n$ applied to an existing feature vector to transform into a new one via scaling.
    
    \item \textit{Transformation}: This operation involves transforming two or more features into a new feature using concatenation, averaging, or more complex fusion techniques, i.e., applying a mapping function on several existing feature vectors via $\mathit{o}_t: \mathbb{R}^n \times \mathbb{R}^m \rightarrow \mathbb{R}^p$, where $n$, $m$, and $p$ are dimensions of the input features and the resulting feature, respectively.
    
    \item \textit{Judgment}: This operation involves logical or rule-based decision-making processes that integrate domain knowledge or empirical rules to form a new feature. In our case, we apply a decision function $\mathit{o}_d : \mathcal{X} \rightarrow \{0, 1\}$, where $\mathcal{X}$ is the input space composed of one or more feature vectors, and the output is a new categorical feature based on predefined criteria.
\end{itemize}

\subsection{Retrieval-Augmented Feature Generation with Reasoning}
In this stage, our objective is to generate a new feature and align it with the existing features in the data table. 
So far, we have obtained external knowledge in terms of the available original features through RAG. The retrieved knowledge contains more structural details and relationships between features, which are not directly reflected in the original dataset. 
To this end, we adopt the LLM to analyze and extract the potential feature structures and relationships, and further generate new features to expand the feature space. 

Given the $k$ documents with potential features $\{g_k\}$, the LLM determines their potential to enhance the performance of downstream tasks via the reasoning capability.
To enable the model to conduct high-quality judgments, we employ a ``Chain of Thought''~\cite{wei2022chain,zhang2024blind,zhang2025ratt} strategy to encourage the LLM to think and formulate logical hypotheses and reasoning for the possible outcomes of integrating each potential feature into the data table step-by-step.
The LLM then selects the most useful documents $\mathcal{R}_t \in \{\mathcal{R}_k\}$ from the candidates, and generates a new feature $g_t$ according to $\mathcal{R}_t$. Finally, we integrate $g_t$ into the data table as 
\begin{equation}
    g_t=\{p_\phi (\mathit{o}(\mathcal{F}_{t-1}) \mid \mathcal{R}_t )\},
    \mathcal{F}_t=\{\mathcal{F}_{t-1},g_t\},
\end{equation}
\begin{equation}
    \mathcal{D}_{t}=\{\mathcal{F}_{t};\mathit{y}\}=\{\{\mathcal{F}_{t-1},g_t\};\mathit{y}\}.
\end{equation}
The $g_t$ is generated by the corresponding combination or judgment method $o_t$ mentioned in document $\mathcal{R}_t$ with existing features and matches the length of $f_i$. Finally, we feed $\mathcal{D}_t$ into the downstream task and obtain the performance metric $\mathcal{P}_t$, which is fed to the LLM for evaluation.

\subsection{Feature Content Validation by LLM}
In this stage, our objective is to validate the effectiveness of the newly generated feature. We decide whether to keep the feature by the improvement of $\mathcal{P}_t- \mathcal{P}_{t-1}$, where a higher value of $\mathcal{P}$ indicates better performance as
\begin{equation}
\mathcal{F}_t = 
\begin{cases} 
\mathcal{F}_t & \text{if } \mathcal{P}_t > \mathcal{P}_{t-1} , \\
\mathcal{F}_{t-1} & \text{otherwise} .
\end{cases}
\end{equation}
If adding the new feature $g_t$ improves downstream task performance, we formally adopt $g_t$ as a new feature within the table. The LLM updates the dataset text information $\mathcal{C}=\mathcal{C}_t=p_\phi(\mathcal{C}_{t-1}|\mathcal{R}_t)$ to incorporate the information related to $g_t$. The feature generation process increases the feature space's dimensionality. As the feature space expands, the model can understand and interpret the data more comprehensively, further improving downstream task performance and enhancing generalization capabilities.

We generate new queries to search for potential features with the updated data table $\mathcal{D}_t$ and the data description $\mathcal{C}_t$. This iteration process continues until a predefined maximum of iterations $T$ is reached, or the optimal feature set is found when achieving peak performance in the downstream task. 
Algorithm \ref{alg:rafg_revised} presents the pseudocode of the proposed method.

\begin{algorithm}[t]
\caption{Retrieval-Augmented Feature Generation}
\label{alg:rafg_revised}
\KwIn{
Tabular dataset $\mathcal{D}_0 = \{\mathcal{F}_0; y\}$, textual information $\mathcal{C}_0$, 
downstream task model $\mathcal{A}$, maximum iterations $T$, patience $K$,
LLM $p_\phi$, performance metric $\mathcal{P}$
}
\KwOut{Enhanced feature set $\mathcal{F}$}

\BlankLine
\textbf{Initialization:}\\
$\mathcal{F} \leftarrow \mathcal{F}_0$, $\mathcal{C} \leftarrow \mathcal{C}_0$, $\mathcal{D} \leftarrow \mathcal{D}_0$, $\text{no\_improve} \leftarrow 0$;\\
$\mathcal{P} \leftarrow \mathcal{A}(\mathcal{D})$; \tcp{Base performance on original features}

\BlankLine
\For{$t = 1$ \KwTo $T$}{
    \tcp{--- Query Generation and Knowledge Retrieval ---}
    $Q_t \leftarrow p_\phi(\mathcal{D}, \mathcal{C})$\;
    Retrieve top-$k$ documents $\{R_k\}$ from knowledge base\;

    \tcp{--- Candidate Evaluation ---}
    \ForEach{$R$ in $\{R_k\}$}{
        $o_k \leftarrow p_\phi(R, \mathcal{F}, \mathcal{C})$\;  \tcp{LLM performs deterministic operation for feature}
        $g_k \leftarrow o_k(\mathcal{F}_{\text{train}})$\;
        $\mathcal{D}_{\text{tmp}} \leftarrow \{\{\mathcal{F}, g_k\}; y\}$\;
        $\mathcal{P}_k \leftarrow \mathcal{A}(\mathcal{D}_{\text{tmp}})$\;
    }

    Select $R_t = \arg\max_k \mathcal{P}_k$ and its corresponding $g_t$\;

    \tcp{--- Validation and Feature Adoption ---}
    \If{$\mathcal{P}_t > \mathcal{P}_{t-1}$}{
        $\mathcal{P} \leftarrow \mathcal{P}_t$; $\mathcal{F} \leftarrow \{\mathcal{F}, g_t\}$\;
        $\mathcal{C} \leftarrow p_\phi(\mathcal{C}\,|\,R_t)$\; \tcp{Update description when feature accepted}
        $\text{no\_improve} \leftarrow 0$\;
    }\Else{
        $\text{no\_improve} \leftarrow \text{no\_improve} + 1$\;
    }

    \If{$\text{no\_improve} \ge K$}{
        \textbf{break}\; \tcp{Early stopping when no improvement for $K$ rounds}
    }
}
\Return $\mathcal{F}$\;
\end{algorithm}

\section{Experiments}

In this section, we present a series of experiments to evaluate the effectiveness and impacts of the \model. We begin by comparing the performance of the \model against several baseline methods on four downstream tasks. To further understand the contributions of different components, we then conduct ablation studies investigating the impact of using an external knowledge library and the choice of different LLMs. Subsequently, we present the information gain observed during the feature generation process. We then showcase the relationship between the newly generated features and the original features, followed by a discussion on the reasons for the observed performance improvements. 

\subsection{Experimental Settings}

\noindent \textbf{Datasets.}
We evaluate the \model on four real-world datasets, including \textit{Parkinson's Disease Classification (PDC)}~\cite{parkinsons_disease_classification_2018}, \textit{Animal Information Dataset (AID)}~\cite{animal_info_dataset_2023}, \textit{Global Country Information Dataset 2023 (GCI)}~\cite{countries_world_2023} and \textit{Diabetes Health Indicators Dataset (DIA)}~\cite{diabetes_health_indicators_dataset_2023}. Table~\ref{tab:dataori} shows the detailed information.

\begin{table}[ht]
\centering
\small
\caption{Datasets from four specific different domains.}

\begin{tabular}{p{1cm}|p{1cm}p{1cm}p{1cm}p{2.5cm}}
    \toprule
    Datasets & Samples & Features & Class & Target \\
    \midrule
    PDC    & 756  & 754    & 2 & Parkinson Binary  \\
    AID    & 205 & 15   & 10 & Social Structure \\
    GCI    & 195  & 34     & 4 & Life Expectancy \\
    DIA    & 253680  & 21    & 2 & Diabetes Binary \\
    \bottomrule
\end{tabular}

\label{tab:dataori}
\vspace{-3mm}
\end{table}

\noindent \textbf{Metrics.} We evaluate the model performance by the following metrics: \textit{Overall Accuracy (Acc)} measures the proportion of true results  
in the total dataset. \textit{Precision (Prec)} reflects the ratio of true positive predictions to all positive predictions for each class. \textit{Recall (Rec)}, also known as sensitivity, reflects the ratio of true positive predictions to all actual positives for each class. \textit{F-Measure (F1)} is the harmonic mean of precision and recall, providing a single score that balances both metrics. We calculate Precision, Recall, and F1-score via macro averaging. 

Aside from evaluating model performance, to further assess the contribution of the features generated by \model, we also utilize \textit{Information Gain (IG)}~\cite{kent1983information}. This metric quantifies the amount of additional information about the target variable \(Y\) that the new features introduce into the dataset. We adopt the concept of information entropy~\cite{wang2022semi2} to calculate IG:
\begin{equation}
   H(Y|F) = -\sum_{y \in Y} p(y|F) \log p(y|F), \label{eq:entropy}
\end{equation}
where \(p(y|F)\) is the probability of class \(y\) given the feature set \(F\). The information gain from expanding an initial feature set \(F_0\) to a new feature set \(F\) is then computed as:
\begin{equation}
   IG(Y; F, F_0) = H(Y|F_0) - H(Y|F), \label{eq:ig}
\end{equation}
which reflects the informational enrichment provided by the generated features~\cite{gao2024information}. Prior studies indicate that features yielding higher information gain are valuable for enhancing predictive model performance~\cite{zhang2024dynamic2}.

\noindent \textbf{Baseline Methods.} We adapt the \model across a range of classification models, including \textit{Random Forests (RF)}~\cite{rigatti2017random}, \textit{Decision Tree (DT)}~\cite{de2013decision}, \textit{TabTransformer (TT)}~\cite{huang2020tabtransformer} and \textit{TabNet (TN)}~\cite{arik2021tabnet}. We compare the performance of each model with and without the features generated by \model. Our \model is also compared against several baselines: the raw data (Raw), the Least Absolute Shrinkage and Selection Operator (Lasso)~\cite{hassan2023comparative}, the Feature Engineering for Predictive Modeling using Reinforcement Learning~\cite{khurana2018feature} (RL), and the Context-Aware Automated Feature Engineering (CAAFE)~\cite{hollmann2024large}.

\noindent \textbf{Implementation Details.} Experiments are conducted on an NVIDIA RTX 4090 GPU. In our implementation for feature generation, we use GPT-4o~\cite{hurst2024gpt} as the query generator paired with Google's Custom Search API to perform the retrieval tasks. 
We use the top-3 results as the retrieved knowledge.
GPT-4o then uses these top relevant results for feature generation. 
For the compared baselines, Lasso and RL are implemented based on standard libraries (scikit-learn~\cite{pedregosa2011scikit} and transformers~\cite{wolf2020transformers}) with hyperparameters tuned via 5-fold cross-validation. CAAFE is configured with default setting with GPT-4o as the generator, and sets the iterations to $3$.

\subsection{Experimental Results}
\noindent \textbf{Overall Performance.} Table~\ref{tab:overall} shows the overall results across four domain-specific datasets. All overall performance results are based on the average of 5-fold cross-validation.
\begin{table*}[ht]
  \centering
    \caption{Overall performance on various downstream tasks across different models. The best and second-best performance is set in \textbf{bold} and \underline{underline}.}
  \resizebox{\textwidth}{!}{
            
        \begin{tabular}{|c|l|cccc|cccc|cccc|cccc|}
    \hline
    \multirow{3}[2]{*}{\textbf{Metrics}} & \multicolumn{1}{c|}{\multirow{3}[2]{*}{\textbf{Model }}} & \multicolumn{4}{c|}{\multirow{3}[2]{*}{\textbf{RF }}} & \multicolumn{4}{c|}{\multirow{3}[2]{*}{\textbf{DT }}} & \multicolumn{4}{c|}{\multirow{3}[2]{*}{\textbf{TT}}} & \multicolumn{4}{c|}{\multirow{3}[2]{*}{\textbf{TN}}} \\
          &       & \multicolumn{4}{c|}{}         & \multicolumn{4}{c|}{}         & \multicolumn{4}{c|}{}         & \multicolumn{4}{c|}{} \\
          &       & \multicolumn{4}{c|}{}         & \multicolumn{4}{c|}{}         & \multicolumn{4}{c|}{}         & \multicolumn{4}{c|}{} \\
    \hline
    \multirow{6}[12]{*}[4ex]{Acc} & Dataset \rule{0pt}{8pt} & PDC   & AID   & GCI   & DIA   & PDC   & AID   & GCI   & DIA   & PDC   & AID   & GCI   & DIA   & PDC   & AID   & GCI   & DIA \\
\cline{2-18}          & Raw \rule{0pt}{8pt}  & 0.820 & 0.635 & 0.674 & 0.759 & 0.773 & 0.423 & 0.571 & 0.794 & 0.841 & 0.577 & 0.306 & 0.861 & 0.677 & 0.365 & 0.510 & 0.859 \\
\cline{2-2}          & Lasso \rule{0pt}{8pt} & 0.847 & 0.654 & 0.735 & 0.860 & 0.788 & 0.442 & 0.592 & 0.796 & 0.857 & 0.615 & 0.327 & 0.862 & 0.709 & 0.423 & 0.531 & 0.860 \\
\cline{2-2}          & RL \rule{0pt}{8pt}    & 0.857 & \underline{0.687} & 0.755 & 0.859 & 0.799 & 0.519 & 0.592 & 0.796 & 0.878 & 0.596 & 0.347 & 0.862 & 0.714 & 0.558 & 0.551 & 0.863 \\
\cline{2-2}          & CAAFE \rule{0pt}{8pt} & \underline{0.862} & 0.673 & \underline{0.776} & \underline{0.863} & \underline{0.820} & \underline{0.577} & \underline{0.612} & \underline{0.797} & \underline{0.884} & \underline{0.654} & \underline{0.408} & \underline{0.863} & \underline{0.741} & \underline{0.577} & \underline{0.592} & \underline{0.865}
 \\
\cline{2-2}          & \textbf{\model} \rule{0pt}{8pt} & \textbf{0.888} & \textbf{0.750} & \textbf{0.846} & \textbf{0.887} & \textbf{0.841} & \textbf{0.615} & \textbf{0.674} & \textbf{0.815} & \textbf{0.900} & \textbf{0.673} & \textbf{0.429} & \textbf{0.865} & \textbf{0.773} & \textbf{0.596} & \textbf{0.714} & \textbf{0.867} \\
    \hline
    \multirow{6}[12]{*}[4ex]{Prec} & Dataset \rule{0pt}{8pt} & PDC   & AID   & GCI   & DIA   & PDC   & AID   & GCI   & DIA   & PDC   & AID   & GCI   & DIA   & PDC   & AID   & GCI   & DIA \\
\cline{2-18}          & Raw \rule{0pt}{8pt}  & 0.721 & 0.175 & 0.675 & 0.681 & 0.688 & 0.118 & 0.551 & 0.588 & 0.797 & 0.218 & 0.327 & 0.575 & 0.490 & 0.106 & 0.474 & 0.429 \\
\cline{2-2}          & Lasso \rule{0pt}{8pt} & 0.753 & 0.188 & 0.690 & 0.684 & 0.755 & 0.150 & 0.607 & 0.590 & 0.797 & 0.237 & 0.307 & \textbf{0.599} & 0.519 & 0.129 & 0.525 & 0.684 \\
\cline{2-2}          & RL \rule{0pt}{8pt}   & \underline{0.760} & 0.245 & 0.763 & 0.683 & 0.722 & \textbf{0.319} & 0.554 & 0.590 & 0.832 & 0.252 & 0.360 & \underline{0.585} & 0.489 & 0.151 & \underline{0.636} & \textbf{0.806} \\
\cline{2-2}          & CAAFE \rule{0pt}{8pt} & 0.753 & \underline{0.237} & \underline{0.768} & \underline{0.724} & \textbf{0.783} & 0.173 & \underline{0.607} & \underline{0.592} & \underline{0.832} & \underline{0.321} & \underline{0.409} & 0.572 & \underline{0.536} & \underline{0.196} & 0.601 & 0.715 \\
\cline{2-2}          & \textbf{\model} \rule{0pt}{8pt} & \textbf{0.793} & \textbf{0.343} & \textbf{0.791} & \textbf{0.735} & \underline{0.777} & \underline{0.261} & \textbf{0.669} & \textbf{0.605} & \textbf{0.857} & \textbf{0.349} & \textbf{0.450} & 0.570 & \textbf{0.559} & \textbf{0.242} & \textbf{0.683} & \underline{0.723} \\
    \hline
    \multirow{6}[12]{*}[4ex]{Rec} & Dataset \rule{0pt}{8pt} & PDC   & AID   & GCI   & DIA   & PDC   & AID   & GCI   & DIA   & PDC   & AID   & GCI   & DIA   & PDC   & AID   & GCI   & DIA \\
\cline{2-18}          & Raw \rule{0pt}{8pt}  & 0.855 & 0.156 & 0.640 & 0.570 & 0.737 & 0.126 & 0.564 & 0.597 & 0.790 & 0.201 & 0.311 & 0.702 & 0.430 & 0.113 & 0.474 & 0.500 \\
\cline{2-2}          & Lasso & 0.818 & 0.194 & 0.720 & 0.575 & 0.727 & 0.141 & 0.594 & \underline{0.599} & 0.820 & 0.224 & 0.333 & 0.696 & 0.583 & 0.101 & 0.533 & 0.538 \\
\cline{2-2}          & RL  \rule{0pt}{8pt}  & 0.842 & 0.206 & 0.742 & 0.568 & 0.742 & 0.244 & 0.549 & 0.598 & \underline{0.846} & 0.236 & \underline{0.373} & 0.691 & 0.440 & 0.101 & 0.533 & 0.501 \\
\cline{2-2}          & CAAFE \rule{0pt}{8pt} & \underline{0.890} & \underline{0.337} & \underline{0.766} & \underline{0.559} & \underline{0.763} & \underline{0.262} & \underline{0.603} & \textbf{0.603}
 & 0.832 & \textbf{0.433} & 0.310 & \underline{0.705} & \underline{0.653} & \underline{0.116} & \underline{0.592} & \underline{0.575}
 \\
\cline{2-2}          & \textbf{\model} \rule{0pt}{8pt} & \textbf{0.905} & \textbf{0.357} & \textbf{0.791} & \textbf{0.563} & \textbf{0.786} & \textbf{0.320} & \textbf{0.703} & 0.596 & \textbf{0.882} & \underline{0.353} & \textbf{0.404} & \textbf{0.710} & \textbf{0.664} & \textbf{0.213} & \textbf{0.669} & \textbf{0.579} \\
    \hline
    \multirow{6}[12]{*}[4ex]{F1} & Dataset \rule{0pt}{8pt} & PDC   & AID   & GCI   & DIA   & PDC   & AID   & GCI   & DIA   & PDC   & AID   & GCI   & DIA   & PDC   & AID   & GCI   & DIA \\
\cline{2-18}          & Raw \rule{0pt}{8pt}  & 0.750 & 0.146 & 0.636 & 0.587 & 0.702 & 0.122 & 0.556 & 0.592 & 0.793 & 0.209 & 0.289 & 0.593 & 0.419 & 0.110 & 0.470 & 0.462 \\
\cline{2-2}          & Lasso & 0.776 & 0.180 & 0.679 & 0.594 & 0.738 & 0.137 & \underline{0.599} & 0.594 & 0.808 & 0.228 & 0.303 & \textbf{0.622} & 0.476 & 0.105 & 0.526 & 0.539 \\
\cline{2-2}          & RL \rule{0pt}{8pt}   & 0.788 & 0.209 & 0.741 & 0.584 & 0.731 & \underline{0.268} & 0.549 & 0.594 & \underline{0.838} & 0.225 & \underline{0.363} & 0.606 & 0.434 & 0.121 & 0.551 & 0.466 \\
\cline{2-2}          & CAAFE \rule{0pt}{8pt} & \underline{0.790} & \underline{0.243} & \underline{0.758} & \underline{0.572} & \underline{0.772} & 0.188 & 0.601 & \underline{0.597} & 0.832 & \textbf{0.344} & 0.350 & \underline{0.589} & \underline{0.508} & \underline{0.137} & \underline{0.579} & \underline{0.595} \\
\cline{2-2}          & \textbf{\model} \rule{0pt}{8pt} & \textbf{0.829} & \textbf{0.324} & \textbf{0.788} & \textbf{0.578} & \textbf{0.781} & \textbf{0.271} & \textbf{0.681} & \textbf{0.600} & \textbf{0.868} & \underline{0.341} & \textbf{0.399} & 0.587 & \textbf{0.556} & \textbf{0.224} & \textbf{0.664} & \textbf{0.600} \\
    \hline
    \end{tabular}
    }
  
\label{tab:overall}
\vspace{-2mm}
\end{table*}
Compared with baseline methods, \model consistently achieves superior performance. For instance, in RF, \model shows a significant improvement in accuracy across all datasets, with a $19.2\%$ increase on AID over raw data, and $3.8\%$ over CAAFE, the second-highest accuracy baseline. The \model shows particular strength in boosting model accuracy. The performance demonstrates that \model's enriching of feature space with domain-specific knowledge enhances the model's understanding and boosts overall performance.

Across different models and metrics, \model consistently outperforms other methods in all metrics. Notably, in GCI dataset under DT, \model achieves a precision of $66.9\%$, surpassing the highest precision of $60.7\%$. Moreover, for F1 scores, \model raises the score in the GCI under DT from $55.6\%$ to $68.1\%$. These results demonstrate the \model's capability of reducing misclassification effectively.

The generation of new features through \model significantly enhances model performance. As results show, adding features, e.g., from 756 to 770 in the PDC dataset, leads to substantial increases in performance metrics, including accuracy and F1. This result suggests that the new features not only numerically expand the feature set, but also enhance the information richness of the dataset. Thus, models can capture more complex patterns and relationships within the data.

\begin{table}[ht]
\centering
\small
\caption{Variation in the number of features.}
\begin{tabular}{p{1cm}|p{0.5cm}>{\raggedleft\arraybackslash}p{1.5cm}|p{1cm}|p{0.5cm}>{\raggedleft\arraybackslash}p{1.5cm}}
    \toprule
    Dataset & Original & \model & Dataset & Original & \model \\
    \midrule
    PDC    & 756    & 770 \textcolor{blue}{(+14)} & GCI    & 34    & 39 \textcolor{blue}{(+5)} \\
    AID    & 15    & 16 \textcolor{blue}{(+1)} & DIA    & 21    & 23 \textcolor{blue}{(+2)} \\
    \bottomrule
\end{tabular}

\label{tab:datanew}
\end{table}



\begin{figure}[htbp]
  \centering
  \includegraphics[width=\columnwidth]{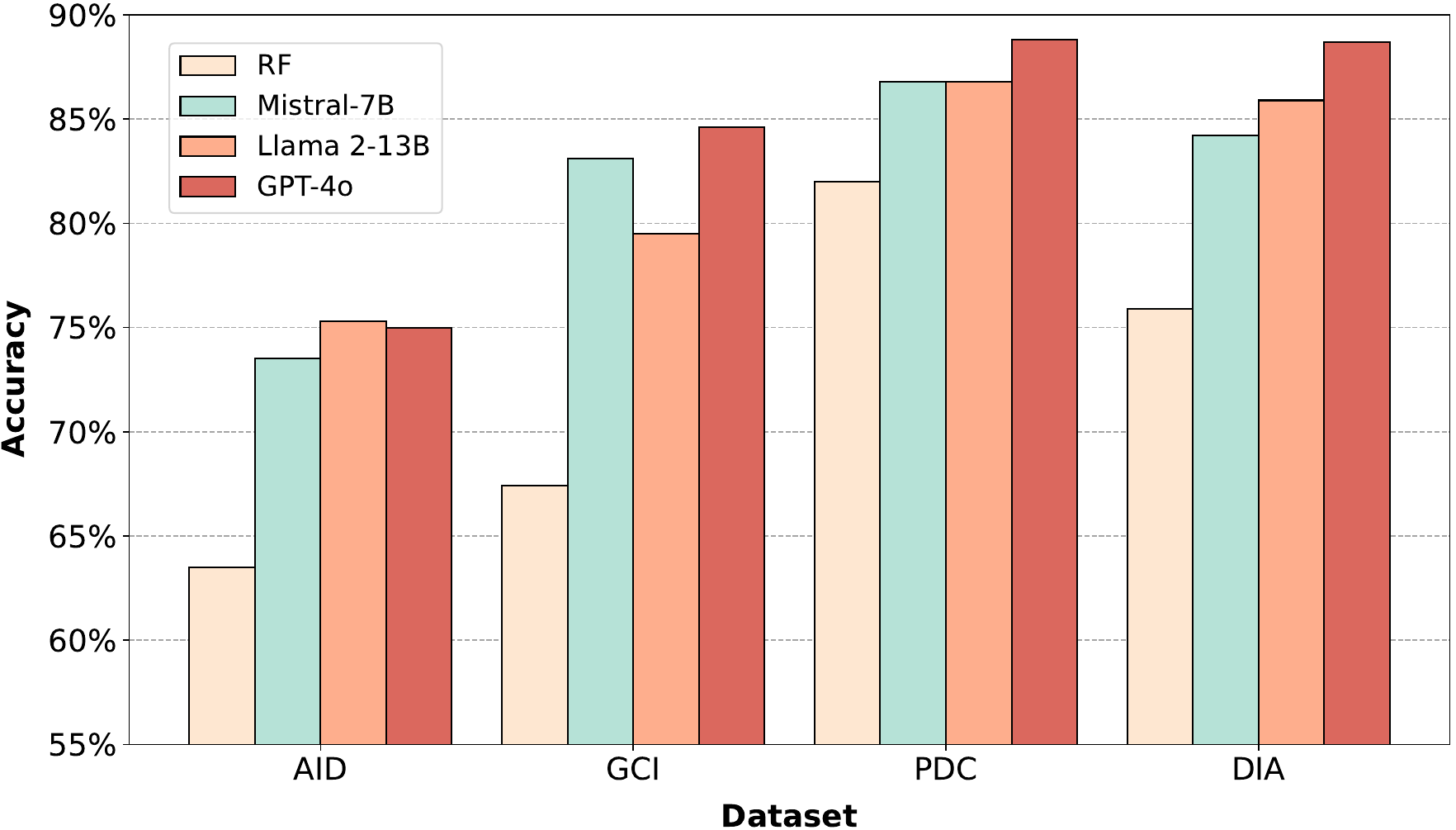} 
  \caption{Accuracy of using RF with \model and various LLMs.} 
  \label{fig:llm_comparison_chart}
  \vspace{-4mm}
\end{figure}

We further assessed the robustness of \model across diverse LLMs: GPT-4o~\cite{hurst2024gpt}, Llama 2-13B~\cite{touvron2023llama}, and Mistral-7B~\cite{jiang2023mistral7b}. As shown in Fig.~\ref{fig:llm_comparison_chart}, integrating \model consistently improves the classification accuracy of the RF classifier on all datasets compared with the Raw baseline. For example, on the PDC dataset, accuracy rises from 0.820 (Raw) to 0.888 with GPT-4o, and similar gains are observed with Llama 2-13B (0.868) and Mistral-7B (0.868). The GCI dataset shows the largest relative improvement, increasing from 0.674 to 0.846 with GPT-4o.  Although GPT-4o achieves the highest overall performance, which is likely because of its stronger reasoning and generation capability,
all three LLMs substantially outperform the baseline when combined with \model.
This improvement indicates that \model possesses good generalizability across LLMs of varying scales and architectures.

\begin{table*}[htbp]
  \centering
  \caption{Some case studies about new features generated by \model based on the GCI dataset.}
  \renewcommand{\arraystretch}{1.1}
  \begin{tabularx}{\linewidth}{|>{\hsize=0.10\hsize}X|>{\hsize=0.85\hsize}X|}
    \hline
    \textbf{Label} & \textbf{Population Load Ratio} \\
    \hline
    \textbf{Calculation} & $$\frac{\text{Population}}{\text{Land Area (Km}^2)}$$ \\
    \hline
    \textbf{Reasoning} &
\makebox[0.97\linewidth][l]{%
  \colorbox{bgcolor2}{%
    \parbox[t]{0.85\linewidth}{%
      \fontsize{10pt}{10pt}\selectfont
      \textit{Represents population density per square kilometer, indicating the population load of a country.}%
    }%
  }%
}
 \\
    \hline
    \textbf{Label} & \textbf{Resource Utilization Rate} \\
    \hline
    \textbf{Calculation} & $$\frac{\text{Agricultural Land (\%)} + \text{Forested Area (\%)}}{100}$$ \\
    \hline
    \textbf{Reasoning} &
\makebox[0.97\linewidth][l]{%
  \colorbox{bgcolor2}{%
    \parbox[t]{0.85\linewidth}{%
      \fontsize{10pt}{10pt}\selectfont
      \textit{Measures the proportion of land used for agriculture and forestry; higher values indicate greater resource use.}
    }%
  }%
}
 \\
    \hline
    \textbf{Label} & \textbf{Education Investment Effectiveness} \\
    \hline
    \textbf{Calculation} & $$\frac{\text{Gross Primary Enrollment (\%)} + \text{Gross Tertiary Enrollment (\%)}}{2}$$ \\
    \hline
    \textbf{Reasoning} &
\makebox[0.97\linewidth][l]{%
  \colorbox{bgcolor2}{%
    \parbox[t]{0.85\linewidth}{%
      \fontsize{10pt}{10pt}\selectfont
      \textit{Reflects investment in primary and higher education, typically linked to better economic development.}
    }%
  }%
}
 \\
    \hline
    \textbf{Label} & \textbf{Environmental Stress Index} \\
    \hline
    \textbf{Calculation} & $$\frac{\text{CO}_2\ \text{Emissions}}{\left( \dfrac{\text{Forested Area (\%)}}{100} \times \text{Land Area (Km}^2) \right)}$$ \\
    \hline
    \textbf{Reasoning} &
\makebox[0.97\linewidth][l]{%
  \colorbox{bgcolor2}{%
    \parbox[t]{0.85\linewidth}{%
      \fontsize{10pt}{10pt}\selectfont
      \textit{Indicates environmental stress by measuring CO$_2$ emissions per unit of forest area. Higher values show greater pressure.}
    }%
  }%
}
 \\
    \hline
    \textbf{Label} & \textbf{GDP per Capita} \\
    \hline
    \textbf{Calculation} & $$\frac{\text{GDP}}{\text{Population}}$$ \\
    \hline
    \textbf{Reasoning} &
\makebox[0.97\linewidth][l]{%
  \colorbox{bgcolor2}{%
    \parbox[t]{0.85\linewidth}{%
      \fontsize{10pt}{10pt}\selectfont
      \textit{Represents a country’s economic level and standard of living.}
    }%
  }%
}
 \\
    \hline
  \end{tabularx}%
  \label{tab:feature}%
\end{table*}

\begin{figure}[htbp]
  \centering
  \vspace{-2mm}
  \includegraphics[width=\columnwidth]{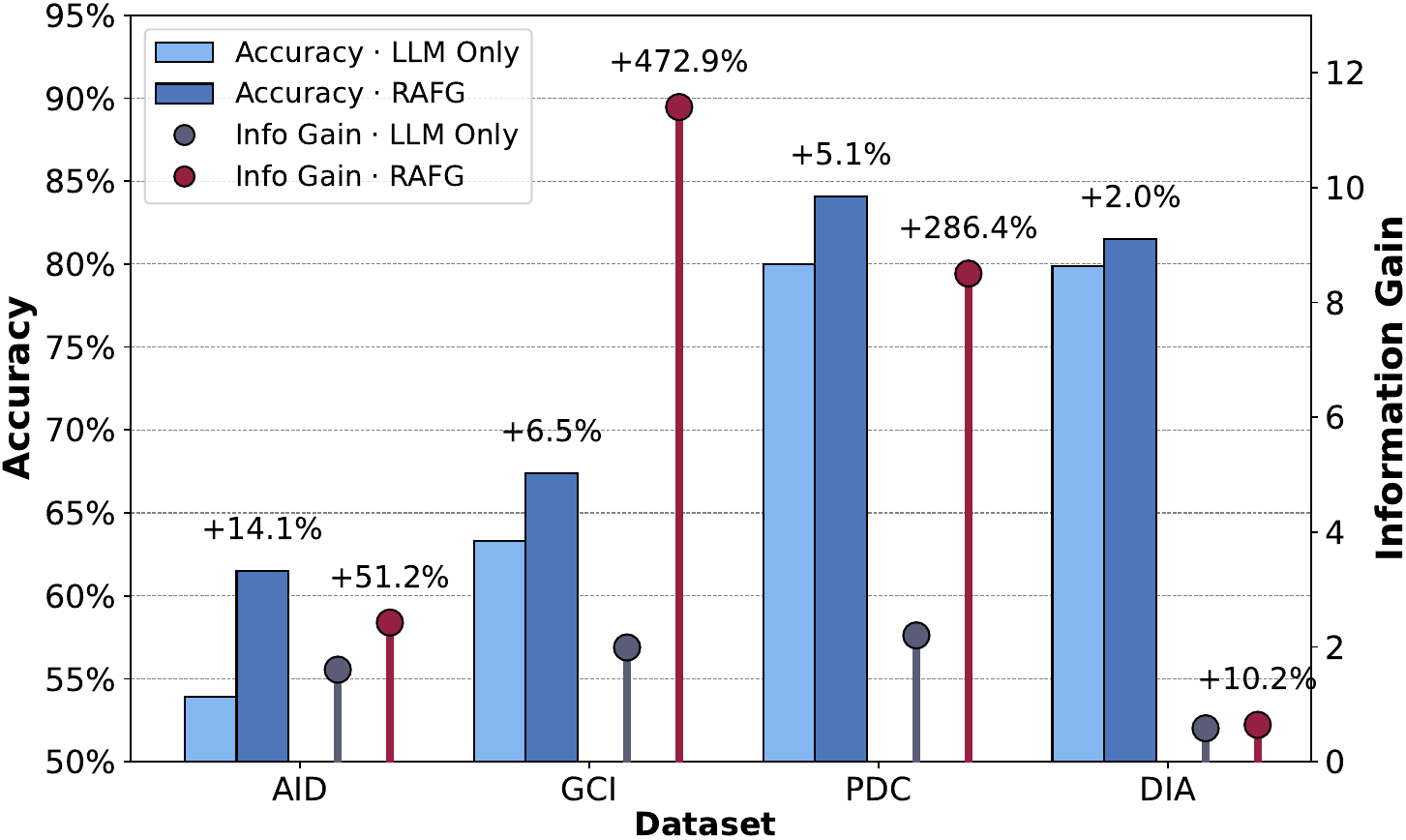}
  \caption{Information gain with \model across different datasets.} 
  \label{fig:informationgain}
  \vspace{-2mm}
\end{figure}

\noindent \textbf{Information Gain.} 
We further study how much addition information is introduced by the new features generated by \model.
We adopt the \textit{Information Gain} (IG) to measure the reduction in uncertainty of the target variable when the generated features are incorporated.
A higher IG value indicates that the new feature provides novel, predictive information about the label's distribution not captured by the original feature set.
As shown in Fig.~\ref{fig:informationgain}, both the prediction accuracy and IG increase consistently with \model. In particular, for PDC and GCI, the IG increases over 200\% and 400\%. This result indicates that the \model-generated features bring new informative signals that make the target variable more predictable for the model. Further, by retrieving external context and generating transformed new features, \model scales the dimension of the feature space for the downstream classifiers to capture the latent structure behind the data.

\begin{table*}[!ht]
\centering
\captionsetup{width=\textwidth}
\caption{Comparison of applying \model on the GCI dataset. The \colorbox{bgcolor1}{blue highlighted text} denotes the new descriptive information generated by the LLM. 
}
\begin{tabular}{@{}llp{\dimexpr\linewidth-4\tabcolsep-2\arrayrulewidth-4cm}@{}}
\toprule
\hline
\textbf{Dataset} & \textbf{Number of Features} & \textbf{Data Description} \\ \midrule
Original & 35 & This comprehensive dataset provides a wealth of information about all countries worldwide, \ldots, enabling in-depth analyses and cross-country comparisons. \\
\hline

 New & 40 \textcolor{blue}{(+5)} & This comprehensive dataset provides a wealth of information about all countries worldwide, \ldots, enabling in-depth analyses and cross-country comparisons. 

\colorbox{bgcolor1}{\parbox{\dimexpr\linewidth-2\fboxsep\relax}{Newly added to this dataset are \textcolor{blue}{five} key variables designed to deepen insights into economic pressures, population density, resource utilization, educational investments, and environmental by stress.}} \\
\hline
\bottomrule
\end{tabular}

\label{tab:dataset_overview}
\vspace{-2mm}
\end{table*}

\noindent \textbf{Case Study.}
To further illustrate how \model generates meaningful and interpretable features, we conduct a detailed case study on the \textit{Global Country Information Dataset 2023 (GCI)}~\cite{countries_world_2023} dataset. This dataset provides well-structured indicators, e.g., economic, demographic, and environmental attributes at the country level. We analyze how the new features contribute to downstream task performance by answering: \textit{a)} what types of features does \model generate and their interpretability; \textit{b)} how do these features quantitatively impact task performance and information gain; and \textit{c)} how do these new features correlate with the original ones and whether they capture more domain-level relations.


\paragraph{Generated Features and Interpretability}
In Table~\ref{tab:feature} and~\ref{tab:dataset_overview} we list five new features generated by \model. Each feature includes a label, a calculation formula, and an automatically generated reasoning explanation. In this case, these features are mainly widely used socio-economic indicators, e.g., \textit{\textbf{the Population Load Ratio}}~\cite{palstra2012effective} integrates \textit{\textbf{Population}} and \textit{\textbf{Land Area}} to describe population density, and \textit{\textbf{Resource Utilization Rate}}~\cite{yu2015improving} integrates \textit{\textbf{Agricultural Land}} and \textit{\textbf{Forested Area}} to describe ecological pressure. These clear and useful generations demonstrate that \model can maintain transparency and semantic meaning based on domain knowledge.

\begin{figure}[ht]
  \centering
  \includegraphics[width=\columnwidth]{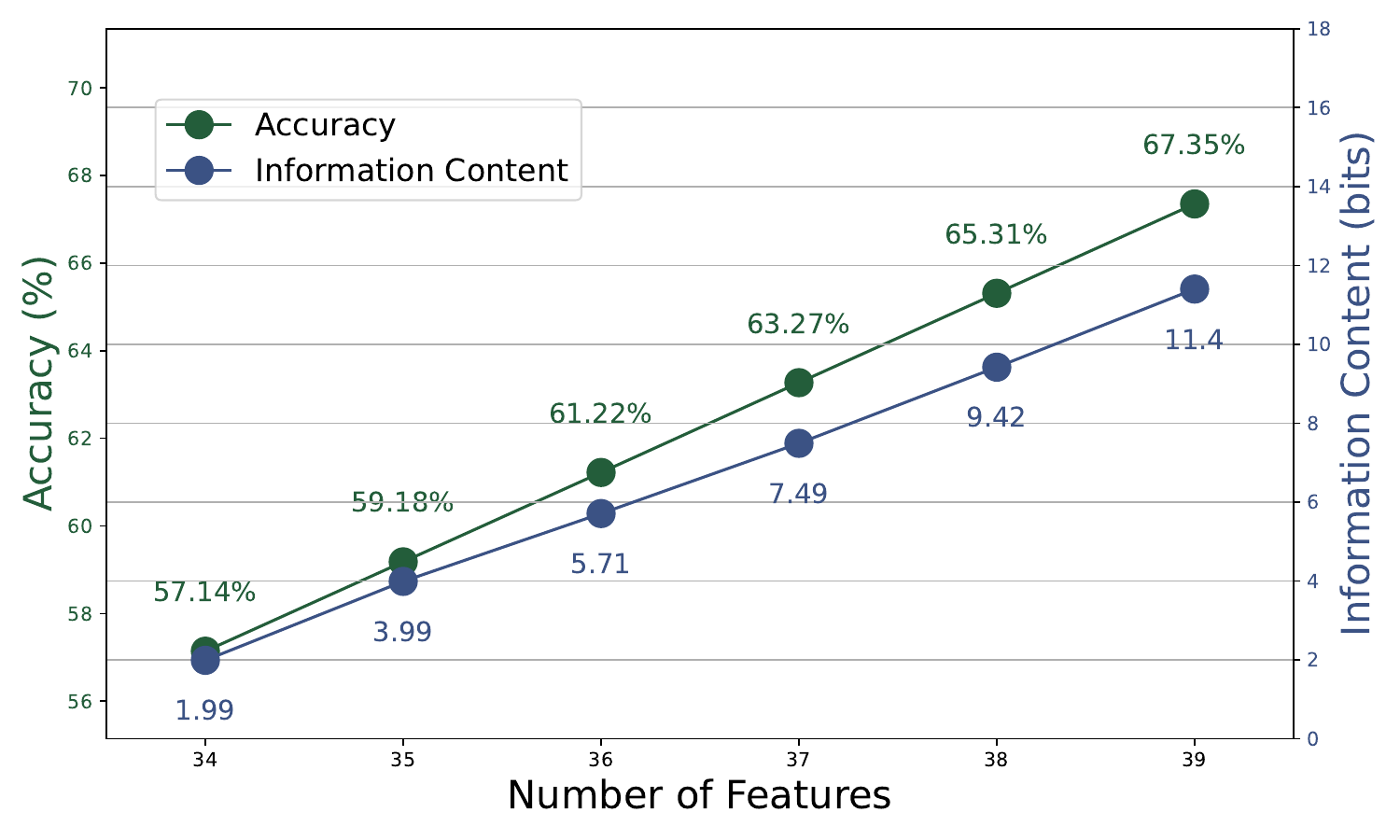}
  \caption{Accuracy variation in \model feature generation process and the information gain for GCI dataset with DT model.}
  \label{fig:acc}
\end{figure}

\begin{figure}[htbp]
  \centering
  \includegraphics[width=\linewidth]{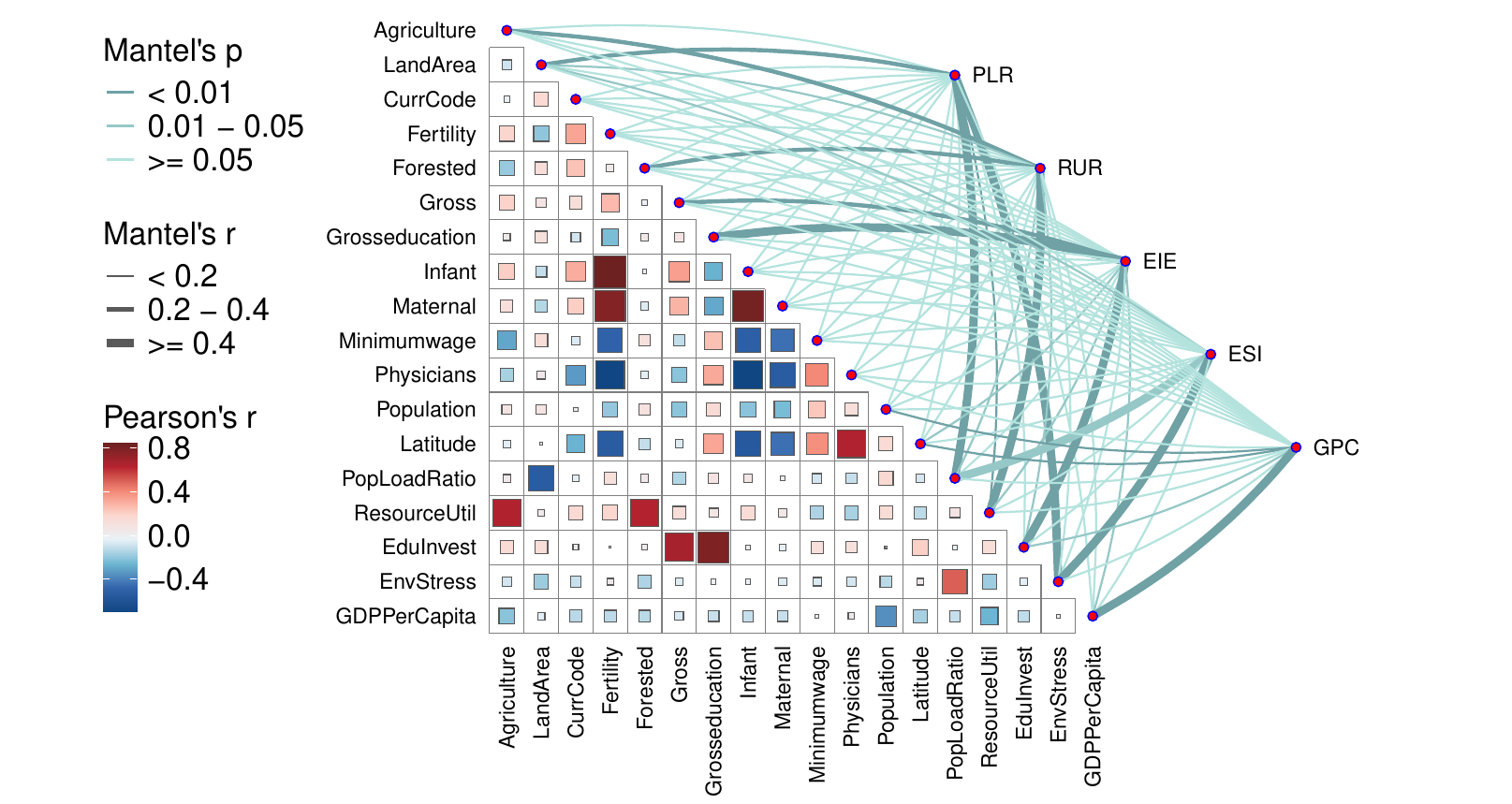}
  \caption{Heatmap of the correlations between the newly generated features (right) and the original features (left).}
  \label{fig:corr}
\vspace{-3mm}
\end{figure}

\paragraph{Performance and Information Gain Evolution}
Fig.~\ref{fig:acc} illustrates how each new feature improves the classification accuracy and increases the dataset's information content. With five new features by \model, the accuracy improves by $10.2\%$ and the IG increases from 1.99 bits to 11.4 bits. This trend demonstrates that each new feature provides distinct and useful information about the target variable.

\paragraph{Correlation} 
To verify that \model does not simply generate redundant transformations, we explore the correlation between the new features generated by \model and the existing features in the GCI dataset. Fig.~\ref{fig:corr} shows a strong positive correlation between these newly generated features and several key features in the dataset. For example, the \textit{\textbf{Population Load Ratio (PLR)}} significantly correlated with geographic indicators like \textit{\textbf{Environmental Stress Index}}, which is because population density interacts with territorial constraints~\cite{ehrlich1971impact,liu2007complexity,shi2020coupling}. Similarly, \textit{\textbf{Resource Utilization Rate (RUR)}} is strongly related to \textit{\textbf{Population Load Ratio}}. This correlation reflects a common real-world pattern in which denser populations cause greater pressure on environmental resources~\cite{dietz2007driving,steffen2015planetary}. In short, the correlations validate that the generated features capture higher-level interactions between the existing factors, which are implicit in the raw data.

\section{Conclusion}

In this paper, we introduce a novel method, \model, that utilizes an LLM for retrieval-augmented feature generation. Our target is to enrich data by effectively utilizing the textual information from external knowledge and LLM's reasoning capability to generate domain-specific features. 
We adopt RAG technique to enhance the correctness and interpretability of feature generation.
The generated features significantly enhance machine learning models without the need for high resource costs or domain-specific fine-tuning. 
The experimental results demonstrate that \model outperforms existing methods in generating useful features and enhancing performance across various domains. Besides, \model's automated and adaptable framework enable the consistent evolves with new data and shows great potential for broader adaptability in various fields. 

\section*{Acknowledgment}

The author Kunpeng Liu is supported by the National Science Foundation (NSF) via the grant numbers 2550105, 2550106, and 2242812.
\begingroup
\bibliographystyle{IEEEtranN}
\bibliography{icdm_2025}

@inproceedings{shuster2021retrieval,
  title={Retrieval Augmentation Reduces Hallucination in Conversation},
  author={Shuster, Kurt and Poff, Spencer and Chen, Moya and Kiela, Douwe and Weston, Jason},
  booktitle={EMNLP},
  pages={3784--3803},
  year={2021}
}

@inproceedings{zamani2022retrieval,
  title={Retrieval-enhanced machine learning},
  author={Zamani, Hamed and Diaz, Fernando and Dehghani, Mostafa and Metzler, Donald and Bendersky, Michael},
  booktitle={SIGIR},
  pages={2875--2886},
  year={2022}
}

@article{lewis2020retrieval,
  title={Retrieval-augmented generation for knowledge-intensive nlp tasks},
  author={Lewis, Patrick and Perez, Ethan and Piktus, Aleksandra and Petroni, Fabio and Karpukhin, Vladimir and Goyal, Naman and K{\"u}ttler, Heinrich and Lewis, Mike and Yih, Wen-tau and Rockt{\"a}schel, Tim and others},
  journal={Advances in neural information processing systems},
  volume={33},
  pages={9459--9474},
  year={2020}
}

@inproceedings{khurana2016cognito,
  title={Cognito: Automated feature engineering for supervised learning},
  author={Khurana, Udayan and Turaga, Deepak and Samulowitz, Horst and Parthasrathy, Srinivasan},
  booktitle={international conference on data mining workshops},
  year={2016},
  organization={IEEE}
}

@article{guo2017deepfm,
  title={DeepFM: a factorization-machine based neural network for CTR prediction},
  author={Guo, Huifeng and Tang, Ruiming and Ye, Yunming and Li, Zhenguo and He, Xiuqiang},
  journal={arXiv preprint arXiv:1703.04247},
  year={2017}
}

@article{yang2024graphusion,
  title={Graphusion: Latent diffusion for graph generation},
  author={Yang, Ling and Huang, Zhilin and Zhang, Zhilong and Liu, Zhongyi and Hong, Shenda and Zhang, Wentao and Yang, Wenming and Cui, Bin and Zhang, Luxia},
  journal={IEEE Transactions on Knowledge and Data Engineering},
  year={2024},
  publisher={IEEE}
}

@article{bjorneld2024real,
  title={Real-world validation of a framework for automated knowledge driven feature engineering inspired by medical domain experts},
  author={Bj{\"o}rneld, Olof and L{\"o}we, Welf},
  journal={Informatics in Medicine Unlocked},
  volume={49},
  pages={101532},
  year={2024},
  publisher={Elsevier}
}

@article{zozoulenko2025random,
  title={Random Feature Representation Boosting},
  author={Zozoulenko, Nikita and Cass, Thomas and Gonon, Lukas},
  journal={arXiv preprint arXiv:2501.18283},
  year={2025}
}

@article{xiao2024traceable,
  title={Traceable group-wise self-optimizing feature transformation learning: A dual optimization perspective},
  author={Xiao, Meng and Wang, Dongjie and Wu, Min and Liu, Kunpeng and Xiong, Hui and Zhou, Yuanchun and Fu, Yanjie},
  journal={ACM Transactions on Knowledge Discovery from Data},
  year={2024},
  publisher={ACM New York, NY}
}

@inproceedings{nargesian2017learning,
  title={Learning Feature Engineering for Classification.},
  author={Nargesian, Fatemeh and Samulowitz, Horst and Khurana, Udayan and Khalil, Elias B and Turaga, Deepak S},
  booktitle={Ijcai},
  volume={17},
  pages={2529--2535},
  year={2017}
}

@article{pan2020deep,
  title={Deep feature generating network: A new method for intelligent fault detection of mechanical systems under class imbalance},
  author={Pan, Tongyang and Chen, Jinglong and Xie, Jingsong and Zhou, Zitong and He, Shuilong},
  journal={IEEE Transactions on Industrial Informatics},
  volume={17},
  number={9},
  pages={6282--6293},
  year={2020},
  publisher={IEEE}
}

@article{wei2022chain,
  title={Chain-of-thought prompting elicits reasoning in large language models},
  author={Wei, Jason and Wang, Xuezhi and Schuurmans, Dale and Bosma, Maarten and Xia, Fei and Chi, Ed and Le, Quoc V and Zhou, Denny and others},
  journal={Advances in neural information processing systems},
  volume={35},
  pages={24824--24837},
  year={2022}
}

@inproceedings{khurana2018feature,
  title={Feature engineering for predictive modeling using reinforcement learning},
  author={Khurana, Udayan and Samulowitz, Horst and Turaga, Deepak},
  booktitle={AAAI},
  volume={32},
  year={2018}
}

@misc{parkinsons_disease_classification_2018,
  author       = {Sakar, C. and Serbes, Gorkem and Gunduz, Aysegul and Nizam, Hatice and Sakar, Betul},
  title        = {Parkinson's Disease Classification},
  year         = {2018},
  howpublished = {UCI Machine Learning Repository}
}

@inproceedings{arik2021tabnet,
  title={Tabnet: Attentive interpretable tabular learning},
  author={Arik, Sercan {\"O} and Pfister, Tomas},
  booktitle={Proceedings of the AAAI conference on artificial intelligence},
  volume={35},
  pages={6679--6687},
  year={2021}
}

@article{rigatti2017random,
  title={Random forest},
  author={Rigatti, Steven J},
  journal={Journal of Insurance Medicine},
  volume={47},
  number={1},
  pages={31--39},
  year={2017},
  publisher={American Academy of Insurance Medicine 1700 Magnavox Way, Fort Wayne, IN 46804}
}

@article{hollmann2024large,
  title={Large language models for automated data science: Introducing caafe for context-aware automated feature engineering},
  author={Hollmann, Noah and M{\"u}ller, Samuel and Hutter, Frank},
  journal={Advances in Neural Information Processing Systems},
  volume={36},
  year={2024}
}

@article{de2013decision,
  title={Decision trees},
  author={De Ville, Barry},
  journal={Wiley Interdisciplinary Reviews: Computational Statistics},
  year={2013},
  publisher={Wiley Online Library}
}

@article{huang2020tabtransformer,
  title={Tabtransformer: Tabular data modeling using contextual embeddings},
  author={Huang, Xin and Khetan, Ashish and Cvitkovic, Milan and Karnin, Zohar},
  journal={arXiv preprint arXiv:2012.06678},
  year={2020}
}

@article{han2023comprehensive,
  title={A comprehensive survey on vector database: Storage and retrieval technique, challenge},
  author={Han, Yikun and Liu, Chunjiang and Wang, Pengfei},
  journal={arXiv preprint arXiv:2310.11703},
  year={2023}
}

@article{mo2025convmix,
  title={ConvMix: A Mixed-Criteria Data Augmentation Framework for Conversational Dense Retrieval},
  author={Mo, Fengran and Zhang, Jinghan and Hui, Yuchen and Sun, Jia Ao and Xu, Zhichao and Su, Zhan and Nie, Jian-Yun},
  journal={arXiv preprint arXiv:2508.04001},
  year={2025}
}

@inproceedings{kaul2017autolearn,
  title={Autolearn—automated feature generation and selection},
  author={Kaul, Ambika and Maheshwary, Saket and Pudi, Vikram},
  booktitle={ICDM},
  year={2017},
  organization={IEEE}
}

@inproceedings{zhang2023openfe,
  title={Openfe: Automated feature generation with expert-level performance},
  author={Zhang, Tianping and Zhang, Zheyu Aqa and Fan, Zhiyuan and Luo, Haoyan and Liu, Fengyuan and Liu, Qian and Cao, Wei and Jian, Li},
  booktitle={International Conference on Machine Learning},
  year={2023},
  organization={PMLR}
}

@article{shi2025flexolmo,
  title={Flexolmo: Open language models for flexible data use},
  author={Shi, Weijia and Bhagia, Akshita and Farhat, Kevin and Muennighoff, Niklas and Walsh, Pete and Morrison, Jacob and Schwenk, Dustin and Longpre, Shayne and Poznanski, Jake and Ettinger, Allyson and others},
  journal={arXiv preprint:2507.07024},
  year={2025}
}

@inproceedings{mo2025uniconv,
  title={UniConv: Unifying Retrieval and Response Generation for Large Language Models in Conversations},
  author={Mo, Fengran and Gao, Yifan and Meng, Chuan and Liu, Xin and Wu, Zhuofeng and Mao, Kelong and Wang, Zhengyang and Chen, Pei and Li, Zheng and Li, Xian and others},
  booktitle={Proceedings of the 63rd Annual Meeting of the Association for Computational Linguistics (Volume 1: Long Papers)},
  pages={6936--6949},
  year={2025}
}

@article{pedregosa2011scikit,
  title={Scikit-learn: Machine learning in Python},
  author={Pedregosa, Fabian and Varoquaux, Ga{\"e}l and Gramfort, Alexandre and Michel, Vincent and Thirion, Bertrand and Grisel, Olivier and Blondel, Mathieu and Prettenhofer, Peter and others},
  journal={the Journal of machine Learning research}
}

@inproceedings{wolf2020transformers,
  title={Transformers: State-of-the-art natural language processing},
  author={Wolf, Thomas and Debut, Lysandre and Sanh, Victor and Chaumond, Julien and Delangue, Clement and Moi, Anthony and Cistac, Pierric and Rault, Tim and Louf, Remi and Funtowicz, Morgan and others},
  booktitle={EMNLP},
  year={2020}
}

@article{yang2024leandojo,
  title={Leandojo: Theorem proving with retrieval-augmented language models},
  author={Yang, Kaiyu and Swope, Aidan and Gu, Alex and Chalamala, Rahul and Song, Peiyang and Yu, Shixing and Godil, Saad and Prenger, Ryan J and Anandkumar, Animashree},
  journal={NIPS},
  year={2024}
}

@article{kent1983information,
  title={Information gain and a general measure of correlation},
  author={Kent, John T},
  journal={Biometrika},
  year={1983},
  publisher={Oxford University Press}
}

@misc{countries_world_2023,
  author = {Fernando Lasso},
  title        = {Countries of the World 2023},
  howpublished = {Kaggle dataset},
}

@misc{animal_info_dataset_2023,
  author       = {Sourav Banerjee},
  title        = {Animal Information},
  howpublished = {Kaggle dataset}
}

@misc{diabetes_health_indicators_dataset_2023,
  title        = {Diabetes Health Indicators},
  author       = {Alex Teboul},
  howpublished = {Kaggle dataset}
}

@article{zhang2024dynamic,
  title={Dynamic and Adaptive Feature Generation with LLM},
  author={Zhang, Xinhao and Zhang, Jinghan and Rekabdar, Banafsheh and Zhou, Yuanchun and Wang, Pengfei and Liu, Kunpeng},
  journal={arXiv preprint arXiv:2406.03505},
  year={2024}
}

@article{zhang2024raft,
  title={Raft: Adapting language model to domain specific rag},
  author={Zhang, Tianjun and Patil, Shishir G and Jain, Naman and Shen, Sheng and Zaharia, Matei and Stoica, Ion and Gonzalez, Joseph E},
  journal={arXiv:2403.10131},
  year={2024}
}

@article{hu2024rag,
  title={Rag and rau: A survey on retrieval-augmented language model in natural language processing},
  author={Hu, Yucheng and Lu, Yuxing},
  journal={arXiv preprint arXiv:2404.19543},
  year={2024}
}

@article{gao2024information,
  title={Information gain ratio-based subfeature grouping empowers particle swarm optimization for feature selection},
  author={Gao, Jinrui and Wang, Ziqian and Jin, Ting and Cheng, Jiujun and Lei, Zhenyu and Gao, Shangce},
  journal={Knowledge-Based Systems},
  volume={286},
  pages={111380},
  year={2024},
  publisher={Elsevier}
}

@article{hassan2023comparative,
  title={A comparative assessment of machine learning algorithms with the Least Absolute Shrinkage and Selection Operator for breast cancer detection and prediction},
  author={Hassan, Md Mehedi and Hassan, Md Mahedi and Yasmin, Farhana and Khan, Md Asif Rakib and Zaman, Sadika and Islam, Khan Kamrul and Bairagi, Anupam Kumar and others},
  journal={Decision Analytics Journal},
  volume={7},
  pages={100245},
  year={2023},
  publisher={Elsevier}

}

@article{han2024large,
  title={Large language models can automatically engineer features for few-shot tabular learning},
  author={Han, Sungwon and Yoon, Jinsung and Arik, Sercan O and Pfister, Tomas},
  journal={arXiv:2404.09491},
  year={2024}
}

@book{dong2018feature,
  title={Feature engineering for machine learning and data analytics},
  author={Dong, Guozhu and Liu, Huan},
  year={2018},
  publisher={CRC press}
}

@article{hurst2024gpt,
  title={Gpt-4o system card},
  author={Hurst, Aaron and Lerer, Adam and Goucher, Adam P and Perelman, Adam and others},
  journal={arXiv preprint:2410.21276},
  year={2024}
}

@inproceedings{li2023learning,
  title={Learning a data-driven policy network for pre-training automated feature engineering},
  author={Li, Liyao and Wang, Haobo and Zha, Liangyu and Huang, Qingyi and Wu, Sai and Chen, Gang and Zhao, Junbo},
  booktitle={The 11th ICLR},
  year={2023}
}

@inproceedings{horn2019autofeat,
  title={The autofeat python library for automated feature engineering and selection},
  author={Horn, Franziska and Pack, Robert and Rieger, Michael},
  booktitle={Joint European Conference on Machine Learning and Knowledge Discovery in Databases},
  year={2019},
  organization={Springer}
}

@article{zhang2025leka,
  title={LEKA: LLM-Enhanced Knowledge Augmentation},
  author={Zhang, Xinhao and Zhang, Jinghan and Mo, Fengran and Wang, Dongjie and Fu, Yanjie and Liu, Kunpeng},
  journal={arXiv preprint arXiv:2501.17802},
  year={2025}
}

@article{zhang2024prototypical,
  title={Prototypical reward network for data-efficient rlhf},
  author={Zhang, Jinghan and Wang, Xiting and Jin, Yiqiao and Chen, Changyu and Zhang, Xinhao and Liu, Kunpeng},
  journal={arXiv preprint arXiv:2406.06606},
  year={2024}
}

@article{wang2025diversity,
  title={Diversity-oriented data augmentation with large language models},
  author={Wang, Zaitian and Zhang, Jinghan and Zhang, Xinhao and Liu, Kunpeng and Wang, Pengfei and Zhou, Yuanchun},
  journal={arXiv preprint arXiv:2502.11671},
  year={2025}
}

@article{wang2022semi2,
  title={Semi-supervised learning for k-dependence Bayesian classifiers},
  author={Wang, LiMin and Zhang, XinHao and Li, Kuo and Zhang, Shuai},
  journal={Applied Intelligence},
  year={2022},
  publisher={Springer}
}

@inproceedings{zhang2024dynamic2,
  title={Dynamic Weight Adjusting Deep Q-Networks for Real-Time Environmental Adaptation},
  author={Zhang, Xinhao and Zhang, Jinghan and Si, Wujun and Liu, Kunpeng},
  booktitle={ICKG},
  year={2024},
  organization={IEEE}
}

@article{zhang2024blind,
  title={Blind Spot Navigation in LLM Reasoning with Thought Space Explorer},
  author={Zhang, Jinghan and Mo, Fengran and Wang, Xiting and Liu, Kunpeng},
  journal={arXiv preprint arXiv:2410.24155},
  year={2024}
}

@inproceedings{zhang2025ratt,
  title={Ratt: A thought structure for coherent and correct llm reasoning},
  author={Zhang, Jinghan and Wang, Xiting and Ren, Weijieying and Jiang, Lu and Wang, Dongjie and Liu, Kunpeng},
  booktitle={Proceedings of the AAAI Conference on Artificial Intelligence},
  year={2025}
}

@inproceedings{zhang2019job2vec,
  title={Job2Vec: Job title benchmarking with collective multi-view representation learning},
  author={Zhang, Denghui and Liu, Junming and Zhu, Hengshu and Liu, Yanchi and Wang, Lichen and Wang, Pengyang and Xiong, Hui},
  booktitle={Proceedings of the 28th ACM International Conference on Information and Knowledge Management},
  pages={2763--2771},
  year={2019}
}

@article{palstra2012effective,
  title={Effective/census population size ratio estimation: a compendium and appraisal},
  author={Palstra, Friso P and Fraser, Dylan J},
  journal={Ecology and evolution},
  volume={2},
  number={9},
  pages={2357--2365},
  year={2012},
  publisher={Wiley Online Library}
}

@article{yu2015improving,
  title={Improving resource utilization efficiency in China's mineral resource-based cities: A case study of Chengde, Hebei province},
  author={Yu, Chenjian and Li, Huiquan and Jia, Xiaoping and Li, Qiang},
  journal={Resources, Conservation and Recycling},
  year={2015},
  publisher={Elsevier}
}

@article{ehrlich1971impact,
  title={Impact of population growth},
  author={Ehrlich, Paul R and Holdren, John P},
  journal={Science},
  year={1971},
  publisher={JSTOR}
}

@article{liu2007complexity,
  title={Complexity of coupled human and natural systems},
  author={Liu, Jianguo and Dietz, Thomas and Carpenter, Stephen R and Alberti, Marina and Folke, Carl and Moran, Emilio and Pell, Alice N and Deadman, Peter and Kratz, Timothy and Lubchenco, Jane and others},
  journal={science},
  volume={317},
  number={5844},
  pages={1513--1516},
  year={2007},
  publisher={American Association for the Advancement of Science}
}

@article{shi2020coupling,
  title={Coupling coordination degree measurement and spatiotemporal heterogeneity between economic development and ecological environment----Empirical evidence from tropical and subtropical regions of China},
  author={Shi, Tao and Yang, Shenyan and Zhang, Wei and Zhou, Qian},
  journal={Journal of Cleaner Production},
  volume={244},
  pages={118739},
  year={2020},
  publisher={Elsevier}
}

@article{dietz2007driving,
  title={Driving the human ecological footprint},
  author={Dietz, Thomas and Rosa, Eugene A and York, Richard},
  journal={Frontiers in Ecology and the Environment},
  volume={5},
  number={1},
  pages={13--18},
  year={2007},
  publisher={Wiley Online Library}
}

@article{steffen2015planetary,
  title={Planetary boundaries: Guiding human development on a changing planet},
  author={Steffen, Will and Richardson, Katherine and Rockstr{\"o}m, Johan and Cornell, Sarah E and Fetzer, Ingo and Bennett, Elena M and Biggs, Reinette and Carpenter, Stephen R and De Vries, Wim and De Wit, Cynthia A and others},
  journal={science},
  volume={347},
  number={6223},
  pages={1259855},
  year={2015},
  publisher={American Association for the Advancement of Science}
}

@article{touvron2023llama,
  title={Llama 2: Open foundation and fine-tuned chat models},
  author={Touvron, Hugo and Martin, Louis and Stone, Kevin and Albert, Peter and Almahairi, Amjad and Babaei, Yasmine and Bashlykov, Nikolay and Batra, Soumya and Bhargava, Prajjwal and Bhosale, Shruti and others},
  journal={arXiv preprint arXiv:2307.09288},
  year={2023}
}

@misc{jiang2023mistral7b,
      title={Mistral 7B}, 
      author={Albert Q. Jiang and Alexandre Sablayrolles and Arthur Mensch and Chris Bamford and Devendra Singh Chaplot and Diego de las Casas and Florian Bressand and Gianna Lengyel and Guillaume Lample and Lucile Saulnier and Lélio Renard Lavaud and Marie-Anne Lachaux and Pierre Stock and Teven Le Scao and Thibaut Lavril and Thomas Wang and Timothée Lacroix and William El Sayed},
      year={2023},
}

@article{kupinski1999feature,
  title={Feature selection with limited datasets},
  author={Kupinski, Matthew A and Giger, Maryellen L},
  journal={Medical Physics},
  year={1999},
  publisher={Wiley Online Library}
}

@article{jain2018feature,
  title={Feature selection and classification systems for chronic disease prediction: A review},
  author={Jain, Divya and Singh, Vijendra},
  journal={Egyptian Informatics Journal},
  year={2018},
  publisher={Elsevier}
}

@article{taha2022using,
  title={Using feature selection with machine learning for generation of insurance insights},
  author={Taha, Ayman and Cosgrave, Bernard and Mckeever, Susan},
  journal={Applied Sciences},
  year={2022},
  publisher={MDPI}
}

@article{zhang2023interpretable,
  title={Interpretable tabular data generation},
  author={Zhang, Yishuo and Zaidi, Nayyar and Zhou, Jiahui and Li, Gang},
  journal={Knowledge and Information Systems},
  volume={65},
  number={7},
  pages={2935--2963},
  year={2023},
  publisher={Springer}
}

@article{kuznetsov2021interpretable,
  title={Interpretable feature generation in ECG using a variational autoencoder},
  author={Kuznetsov, VV and Moskalenko, VA and Gribanov, DV and Zolotykh, Nikolai Yu},
  journal={Frontiers in genetics},
  year={2021},
  publisher={Frontiers Media SA}
}

@inproceedings{zhang2021domain,
  title={Domain-oriented language modeling with adaptive hybrid masking and optimal transport alignment},
  author={Zhang, Denghui and Yuan, Zixuan and Liu, Yanchi and Liu, Hao and Zhuang, Fuzhen and Xiong, Hui and Chen, Haifeng},
  booktitle={Proceedings of the 27th ACM SIGKDD Conference on Knowledge Discovery \& Data Mining},
  year={2021}
}

@inproceedings{zhang2022learning,
  title={Learning to walk with dual agents for knowledge graph reasoning},
  author={Zhang, Denghui and Yuan, Zixuan and Liu, Hao and Xiong, Hui and others},
  booktitle={Proceedings of the AAAI Conference on artificial intelligence},
  volume={36},
  number={5},
  pages={5932--5941},
  year={2022}
}

@article{jeong2024llm,
  title={Llm-select: Feature selection with large language models},
  author={Jeong, Daniel P and Lipton, Zachary C and Ravikumar, Pradeep},
  journal={arXiv preprint arXiv:2407.02694},
  year={2024}
}

@inproceedings{fan2024survey,
  title={A survey on rag meeting llms: Towards retrieval-augmented large language models},
  author={Fan, Wenqi and Ding, Yujuan and Ning, Liangbo and Wang, Shijie and Li, Hengyun and Yin, Dawei and Chua, Tat-Seng and Li, Qing},
  booktitle={Proceedings of the 30th ACM SIGKDD conference on knowledge discovery and data mining},
  year={2024}
}

@article{wang2024controllable,
  title={Controllable data generation by deep learning: A review},
  author={Wang, Shiyu and Du, Yuanqi and Guo, Xiaojie and Pan, Bo and Qin, Zhaohui and Zhao, Liang},
  journal={ACM Computing Surveys},
  year={2024},
  publisher={ACM New York, NY}
}

@article{islam2022comprehensive,
  title={A comprehensive survey on the process, methods, evaluation, and challenges of feature selection},
  author={Islam, Md Rashedul and Lima, Aklima Akter and Das, Sujoy Chandra and Mridha, Muhammad Firoz and Prodeep, Akibur Rahman and Watanobe, Yutaka},
  journal={IEEE Access},
  year={2022},
  publisher={IEEE}
}

@inproceedings{gong2025evolutionary,
  title={Evolutionary large language model for automated feature transformation},
  author={Gong, Nanxu and Reddy, Chandan K and Ying, Wangyang and Chen, Haifeng and Fu, Yanjie},
  booktitle={AAAI},
  year={2025}
}

@inproceedings{galindo2023large,
  title={Large language models to generate meaningful feature model instances},
  author={Galindo, Jos{\'e} A and Dominguez, Antonio J and White, Jules and Benavides, David},
  booktitle={Proceedings of the 27th ACM International Systems and Software Product Line Conference-Volume A},
  pages={15--26},
  year={2023}
}

@article{yu2023large,
  title={Large language model as attributed training data generator: A tale of diversity and bias},
  author={Yu, Yue and Zhuang, Yuchen and Zhang, Jieyu and Meng, Yu and Ratner, Alexander J and Krishna, Ranjay and Shen, Jiaming and Zhang, Chao},
  journal={NIPS},
  year={2023}
}

@article{yu2024auto,
  title={Auto-rag: Autonomous retrieval-augmented generation for large language models},
  author={Yu, Tian and Zhang, Shaolei and Feng, Yang},
  journal={arXiv preprint arXiv:2411.19443},
  year={2024}
}

@article{tricot2014domain,
  title={Domain-specific knowledge and why teaching generic skills does not work},
  author={Tricot, Andr{\'e} and Sweller, John},
  journal={Educational psychology review},
  year={2014},
  publisher={Springer}
}

@inproceedings{wang2023human,
  title={Human-instructed deep hierarchical generative learning for automated urban planning},
  author={Wang, Dongjie and Wu, Lingfei and Zhang, Denghui and Zhou, Jingbo and Sun, Leilei and Fu, Yanjie},
  booktitle={Proceedings of the AAAI Conference on Artificial Intelligence},
  year={2023}
}

@article{chandrashekar2014survey,
  title={A survey on feature selection methods},
  author={Chandrashekar, Girish and Sahin, Ferat},
  journal={Computers \& electrical engineering},
  year={2014},
  publisher={Elsevier}
}

@inproceedings{salau2019feature,
  title={Feature extraction: a survey of the types, techniques, applications},
  author={Salau, Ayodeji Olalekan and Jain, Shruti},
  booktitle={2019 international conference on signal processing and communication (ICSC)},
  year={2019},
  organization={IEEE}
}

@article{dhal2022comprehensive,
  title={A comprehensive survey on feature selection in the various fields of machine learning},
  author={Dhal, Pradip and Azad, Chandrashekhar},
  journal={Applied intelligence},
  year={2022},
  publisher={Springer}
}

@article{sanmartin2024kg,
  title={Kg-rag: Bridging the gap between knowledge and creativity},
  author={Sanmartin, Diego},
  journal={arXiv:2405.12035},
  year={2024}
}
\endgroup

\end{document}